\documentclass[10pt,twocolumn,letterpaper]{article}

\usepackage{iccv}
\usepackage{times}
\usepackage{epsfig}
\usepackage{graphicx}
\usepackage{amsmath}
\usepackage{amssymb}
\usepackage{bbm}
\usepackage{algorithm}  
\usepackage{algorithmic}
\usepackage{stfloats}
\usepackage{booktabs} 

\usepackage[T1]{fontenc}
\usepackage[utf8]{inputenc}
\usepackage{authblk}
\usepackage{newunicodechar}
\newunicodechar{ﬁ}{fi}
\newunicodechar{ﬂ}{fl}

\usepackage[breaklinks=true,bookmarks=false,colorlinks,urlcolor=black]{hyperref}

\iccvfinalcopy 

\def\iccvPaperID{7410} 

\ificcvfinal\fi

\begin{document}

\title{LeafMask: Towards Greater Accuracy on Leaf Segmentation}

\newcommand*\samethanks[1][\value{footnote}]{\footnotemark[#1]}
\author[1,4]{Ruohao Guo}
\author[1,4]{Liao Qu}
\author[3]{Dantong Niu}
\author[1,4,5]{Zhenbo Li\thanks{Corresponding author. E-mail:
\href{mailto:lizb@cau.edu.cn}{lizb@cau.edu.cn}}}
\author[2]{Jun Yue\samethanks}
\affil[1]{College of Information and Electrical Engineering, China Agricultural University}
\affil[2]{College of Information and Electrical Engineering, LuDong University}
\affil[3]{EECS department, University of California, Berkeley}
\affil[4]{Key Laboratory of Agricultural Information Acquisition Technology, Ministry of Agriculture}
\affil[5]{National Innovation Center for Digital Fishery, China Agricultural University}

\maketitle
\ificcvfinal\thispagestyle{empty}\fi

\begin{abstract}
   Leaf segmentation is the most direct and effective way for high-throughput plant phenotype data analysis and quantitative researches of complex traits. Currently, the primary goal of plant phenotyping is to raise the accuracy of the autonomous phenotypic measurement.
   In this work, we present the LeafMask neural network, a new end-to-end model to delineate each leaf region and count the number of leaves, with two main components: 1) the mask assembly module merging position-sensitive bases of each predicted box after non-maximum suppression (NMS) and corresponding coefficients to generate original masks; 2) the mask refining module elaborating leaf boundaries from the mask assembly module by the point selection strategy and predictor.
   In addition, we also design a novel and flexible multi-scale attention module for the dual attention-guided mask (DAG-Mask) branch to effectively enhance information expression and produce more accurate bases.
   Our main contribution is to generate the final improved masks by combining the mask assembly module with the mask refining module under the anchor-free instance segmentation paradigm.
   We validate our LeafMask through extensive experiments on Leaf Segmentation Challenge (LSC) dataset. Our proposed model achieves the 90.09\% BestDice score outperforming other state-of-the-art approaches.
\end{abstract}

\section{Introduction}
Plant phenotyping is a series of quantitative descriptions and methodologies for the morphological, physiological, genetical, and biochemical characteristics of the plant. Phenotyping is often used to improve agricultural management and select excellent crop breeds based on the desired traits and environments. Traditional phenotypic measurement and analysis are laborious, expensive, destructive, and time-consuming. With the rapid development of non-invasive and digital technologies, image-based phenotyping has become a crucial tool for measuring and accessing the plant’s structural and functional properties in high-throughput phenotyping \cite{li2020review}.

\begin{figure}
\begin{center}
\includegraphics[width=\linewidth]{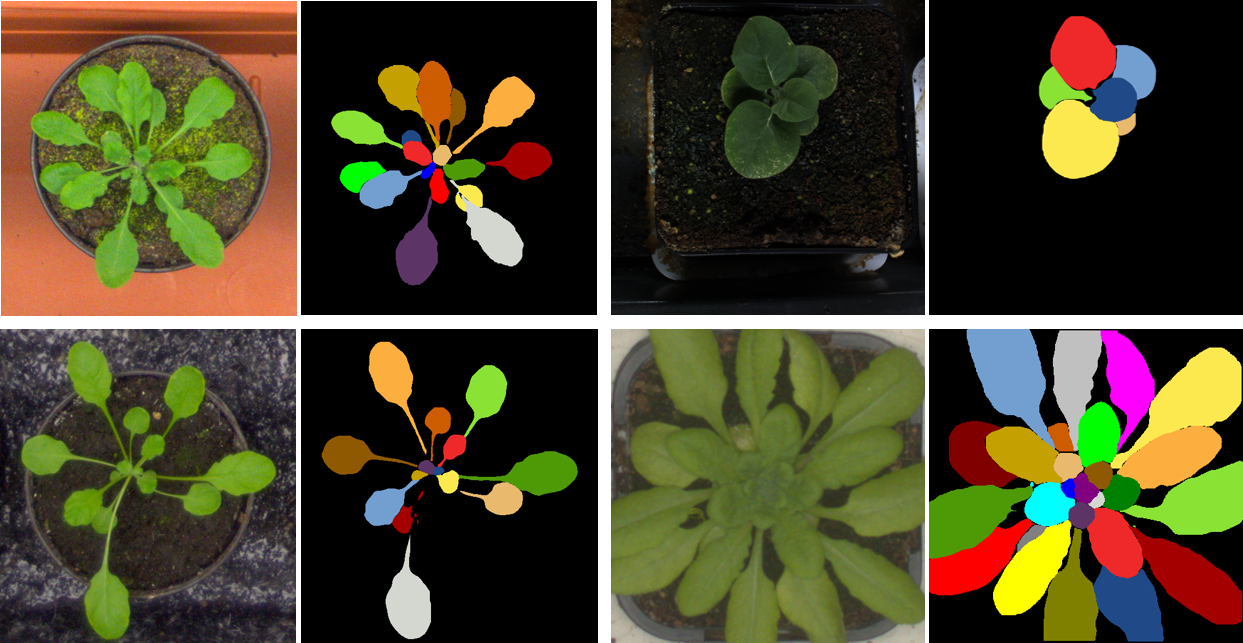}
\end{center}
\vspace{-0.2cm}
   \caption{\textbf{Example images and corresponding predicted masks} of Arabidopsis and Tobacco from the LSC test sets (A1--A4).}
\label{fig1}
\vspace{-1.5em}
\end{figure}

Among all the organs in most plants, leaves account for the largest proportion and play a vital role in the growth and development of vegetation. Leaf properties include but are not limited to leaf count, leaf area, leaf shape characteristics (e.g., leaf length, leaf width, leaf angle, etc.), and leaf nutrient content (e.g., water content, trace element content, etc.). Leaf properties are closely related to many biological and physical processes of plants, such as photosynthesis, respiration, transpiration, and carbon and nutrient cycles \cite{silva2020soil}. Therefore, the estimation of leaf anatomical structure and ontogenetical parameters is of great significance to plant growth monitoring \cite{damian2020natural}. In addition, observation of leaves can also reveal their growth status and ultimately help us to identify genetic contributions, improve plant genetic characteristics, and increase crop yield \cite{mathan2016enhancing}.

In high-throughput phenotyping, the automatic segmentation of plant leaves is a pre-requisite for measuring and identifying more complex phenotypic traits. However, in the perspective of computer vision, obtaining such detailed information at the individual leaf level is particularly difficult \cite{scharr2016leaf}. Despite the clear appearance and shape characteristics, individual leaf segmentation faces the following challenges: the occlusion and overlap of leaves, 
half-covered young leaf and petiole, over-segmentation of the leaf along a visually prominent vein, hard distinct shadowing on the large leaf, reflective leaf surface, as well as the variability in leaf shapes and sizes over the life-cycle of the constantly changing plant \cite{ward2018deep}.

Deep learning is proved to be a powerful tool to segment and count objects, which can avoid the manual design of feature extractors and the laborious selection of parameters. In general, most instance segmentation methods based on convolutional neural networks (CNNs) are split into two branches: object detection and segmentation. The former detects and distinguishes instances (or leaves) to generate bounding boxes, while the latter is employed to separate the foreground and background in the above bounding boxes. According to the object detectors, instance segmentation can be divided into anchor-based methods \cite{bolya2019yolact, he2017mask} and anchor-free methods \cite{2020BlendMask, Lee2020CenterMask}. However, the above methods are used for general computer vision tasks and common object datasets, and their effectiveness in the leaf segmentation task or other agricultural applications needs further exploration. Our main purpose of this paper is to propose an anchor-free method for leaf segmentation.

In this paper, we present a new end-to-end leaf segmentation network called LeafMask (leaf segmentation results are displayed in Figure \ref{fig1}). The innovations of LeafMask are summarized as follows: \textbf{1) Mask assembly module.} The proposed LeafMask can learn to localize and distinguish different leaves via the mask assembly module which merges position-sensitive bases and corresponding predicted coefficients. This framework can effectively avoid some drawbacks of anchor-based methods: \emph{(i)} The hyper-parameters of the anchor boxes (e.g., the sizes, aspect ratios, quantities, etc.) are sensitive and needed to be carefully tuned. \emph{(ii)} Due to the fixed sizes and aspect ratios of anchor boxes, anchor-based methods are difficult to deal with object candidates with large shape variations, especially for small objects. \emph{(iii)} Most of the dense anchor boxes are labelled as negative samples during training, which aggravates the imbalance between positive and negative samples. \textbf{2) DAG-Mask branch.} Since multi-scale attention module aggregates global and local features, it can capture more detailed information and suppress noise of irrelevant clutters. Therefore, it is able to trade off the quality of the model on large blades with that on small leaves. \textbf{3) Mask refining module.} Mask refining module is a simple additional component that adaptively selects points in boundaries of leaves (the most uncertain point set) and efficiently computes sharp boundaries between leaves to avoid aliasing effects. After mask refining, leaf masks are more accurate, particularly in the leaf boundaries. 

To provide evidence for the above innovations, we evaluate LeafMask on leaf segmentation task using the challenging LSC dataset. We carry out extensive ablation studies to discover the optimal hyper-parameters. The performance of our model achieves the 90.09\% BestDice score on the test set and outperforms other state-of-the-art approaches.

\section{Related Work}

\subsection{Leaf segmentation}
Several authors applied context information in time-lapse images to segment and track leaves. For instance, Dellen et al. segmented and tracked the leaves in a set of tobacco-plant growth sequences by graph-based tracking algorithm \cite{dellen2015growth}. Yin et al. used a Chamfer matching formulation to separate and track the leaves of Arabidopsis in the subsequent fluorescence video frames \cite{yin2014multi}. Evidently, the above approaches rely on extra context or temporal information, and therefore are unsuitable for a common reference dataset of individual images. In order to solve this problem, Pape and Klukas utilized 3D histograms of LAB color space to discriminate foreground from background, and then adopted distance maps and region growing algorithm to separate the individual leaves \cite{agapito2015computer}. Another study employed a superpixel-based unsupervised method to extract foreground, and then individual leaves were divided by computing distance maps and using a watershed transform \cite{scharr2016leaf}. In conclusion, the two methods depend on accurate post-processing of distance maps to segment leaves.

\subsection{Instance segmentation}
\textbf{Anchor-based instance segmentation.}
Mask R-CNN \cite{he2017mask} is a representative anchor-based and two-stage instance segmentation approach that first generates a set of candidate Regions of Interests (RoIs) by Region Proposal Network (RPN) on CNN feature maps and then classifies and segments those RoIs in the second stage. Ward et al. trained a Mask R-CNN with a combination of real and synthetic images of rosette plants with different shapes, and used the trained model to segment real leaves \cite{ward2018deep}. In order to predict masks with substantially finer detail around object boundaries, Kirillov et al. proposed a new module called PointRend \cite{kirillov2020pointrend} that viewed image segmentation as a rendering problem and applied a subdivision strategy to adaptively select a non-uniform set of points and efficiently produce more accurate segmentation maps. Compared with Mask R-CNN and PointRend, YOLACT \cite{bolya2019yolact} is an anchor-based but one-stage instance segmentation method, which closely follows a one-stage detector named RetinaNet with an advantage on speed. Instead of using position-controlled tiles or localization steps (e.g., RoI Align), a set of mask coefficients and prototype masks are produced by $fc$ layers and $conv$ layers, respectively. Then this set of coefficients and bottom mask bases can be implemented as a single matrix multiplication to generate the final mask.

\textbf{Anchor-free instance segmentation.}
Recent advances in anchor-free segmentation proved that anchor-free methods can outperform their anchor-based counterparts in accuracy \cite{2020BlendMask}. Instances are freely matched to prediction features without the restrictions of predefined anchor boxes and largely improve the efficiency and precision of segmentation. BlendMask \cite{2020BlendMask} consists of an anchor-free detector network and a mask branch. The framework builds upon the FCOS object detector \cite{Tian2019FCOS} and appends a single convolution layer to predict top-level attentions. Unlike the mask coefficients in YOLACT, which merely performs a weighted sum of the channels of the prototypes, the attention map is a tensor corresponding to the weights at each location of the prototypes. Therefore, BlendMask can provide and encode more instance-level information such as the coarse shape and pose of the object. CenterMask \cite{Lee2020CenterMask} is also a simple but efficient anchor-free instance segmentation, which adds a novel spatial attention-guided branch to predict a segmentation mask on each box.
Our idea is inspired by the above anchor-free instance segmentation methods and LeafMask relies on this segmentation paradigm.

\begin{figure*}
\begin{center}
\includegraphics[width=0.9\linewidth]{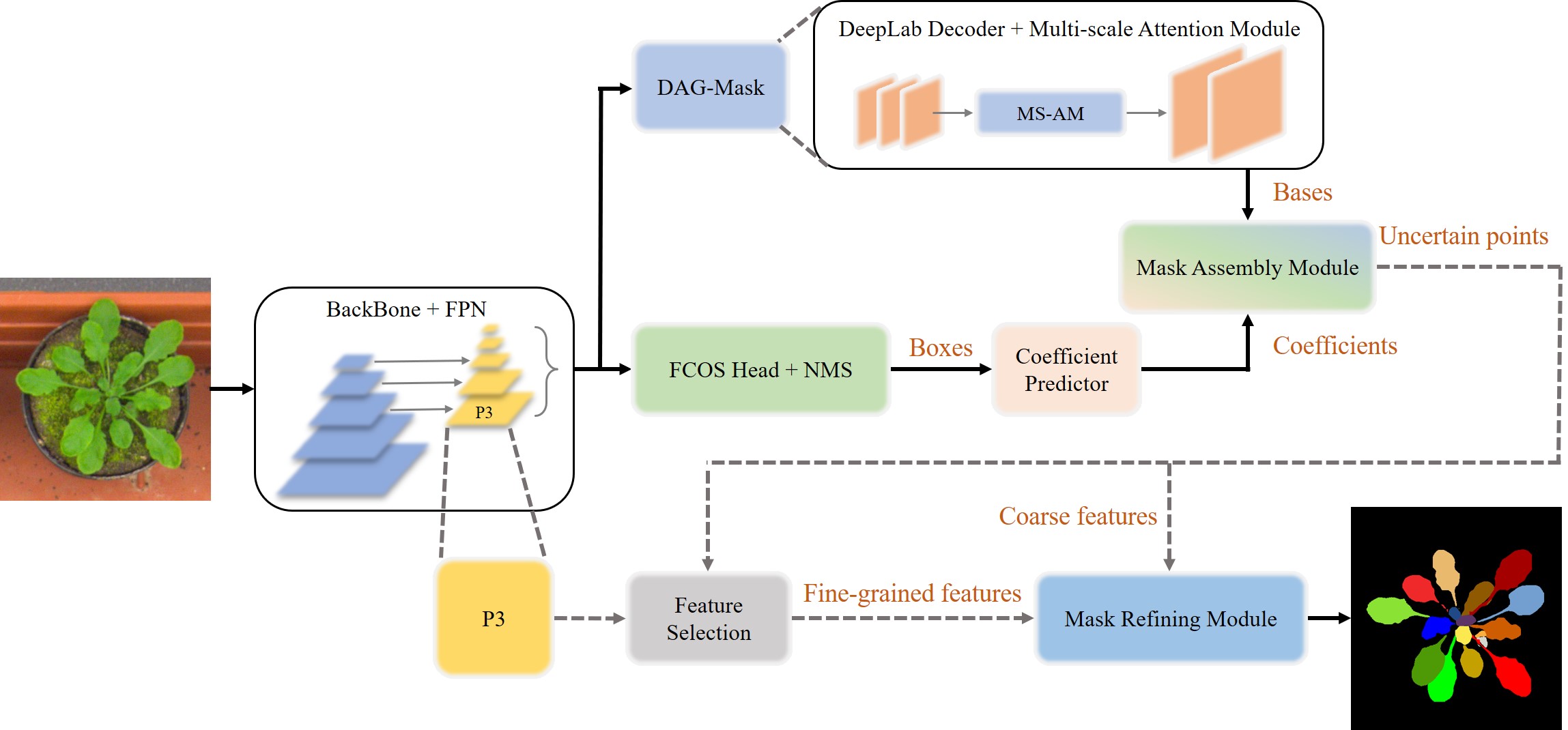}
\end{center}
   \caption{\textbf{LeafMask Architecture.} The backbone and neck of our model adopt ResNet101 and FPN, respectively. Using FPN features, the state-of-the-art FCOS detector predicts bounding boxes including classification, box regression, and centerness. The DAG-Mask branch consists of a new proposed multi-scale attention module and DeepLab V3+ decoder to produce original segmentation masks inside each detected leaf from the shared FCOS head. Mask assembly module linearly combines the outputs of coefficient predictor and DAG-Mask branch (specific assemble process is shown in Figure \ref{fig6}). By selecting a series of uncertain points based on point selection strategy, mask refining module recalculates these points from coarse and fine-grained features to generate the final segmentation image.}
\label{fig2}
\vspace{0em}
\end{figure*}

\subsection{Attention mechanism}
Attention mechanism plays a vital role in various tasks \cite{hu2018squeeze, lin2016efficient, tang2011image}. Attention modules can selectively establish long-range context dependencies and focus on the significant information from a large amount of information to improve the efficiency and accuracy of processing models. SENet \cite{hu2018squeeze} squeezes the channel-wise global spatial information of the input feature map into a channel descriptor to model inter-channel dependencies by using global average pooling. Compared with the SENet, Woo et al. experimentally confirmed that global max pooling can obtain other crucial clues about distinctive object features \cite{woo2018cbam}. Accordingly, using global average-pooled and max-pooled features simultaneously to infer finer channel-wise attention greatly improves the ability of representations produced by a network rather than using each independently.

Traditional attention module discriminates the feature representations for scene-and-object understanding by capturing the global information. However, the global attention only learns one single scalar value for a spatial position or feature map. We consider that using global and coarse descriptors to encode leaves is suboptimal, which can potentially ignore or suppress most of the image signal present in small leaves. As a result, we propose the multi-scale attention module with both spatial and channel descriptions to combine the global and local contexts inside the dual attention-guided mask branch.

\section{Methods}
We separate leaf segmentation into two parallel parts: detection and mask branch, first detecting and generating the leaf bounding boxes, and then predicting the foreground masks on each box. The former adopts the state-of-the-art FCOS object detector with minimal parameter modification. The latter mainly includes three parts: dual attention-guided mask (DAG-Mask) branch, mask assembly module, and mask refining module (Figure \ref{fig2}).

\subsection{DAG-Mask branch}
\begin{figure}[hp]
\begin{center}
\includegraphics[width=\linewidth]{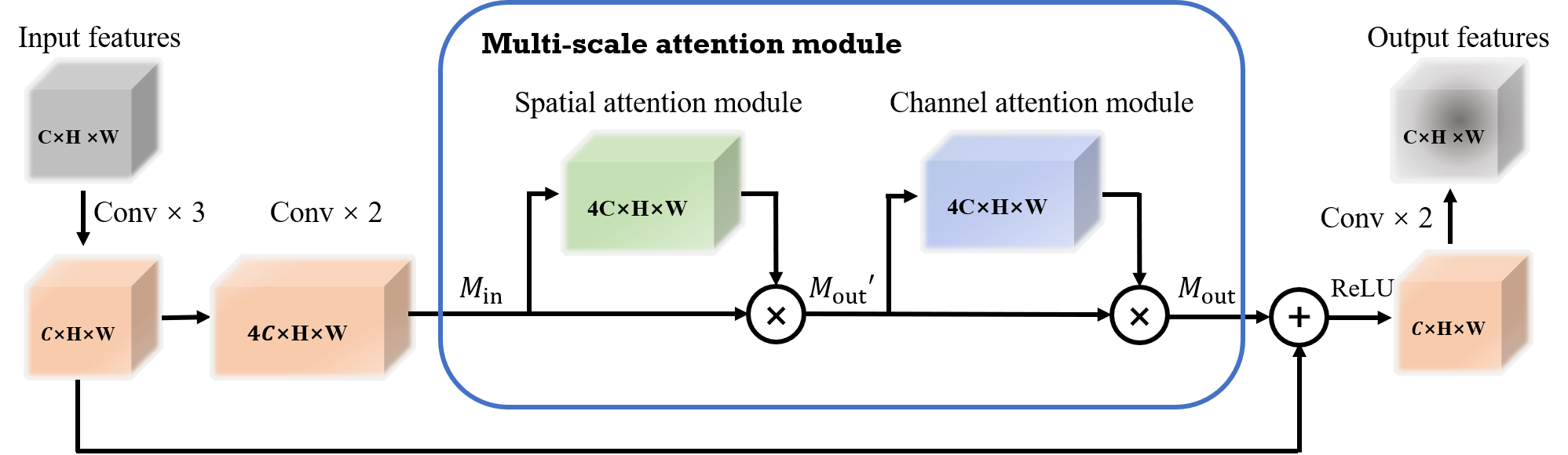}
\end{center}
\vspace{-0.3cm}
   \caption{\textbf{Diagram of the proposed DAG-Mask branch}, where $\bigoplus$ denotes the element-wise addition and $\bigotimes$ denotes the element-wise multiplication.}
\label{fig3}
\vspace{-0.5em}
\end{figure}

Similar to other convolutional instance-aware segmentation methods \cite{bolya2019yolact, 2020BlendMask}, we add a bottom module and use the decoder of DeepLab V3+ to generate a set of non-local score maps (i.e., bases) over the entire image. To better boost the representation power of DeepLab V3+ and focus on target objects properly, we propose a new dual attention module with spatial and channel descriptions, which aggregates global and local features in DAG-Mask branch.

Given the feature pyramid map as input, we first feed it into convolutional layers to expand channels and produce a new feature map. Then, we apply the multi-scale attention module to sequentially infer spatial attention map and channel attention map as illustrated in Figure \ref{fig3}.

\begin{figure}[hp]
\begin{center}
\includegraphics[width=\linewidth]{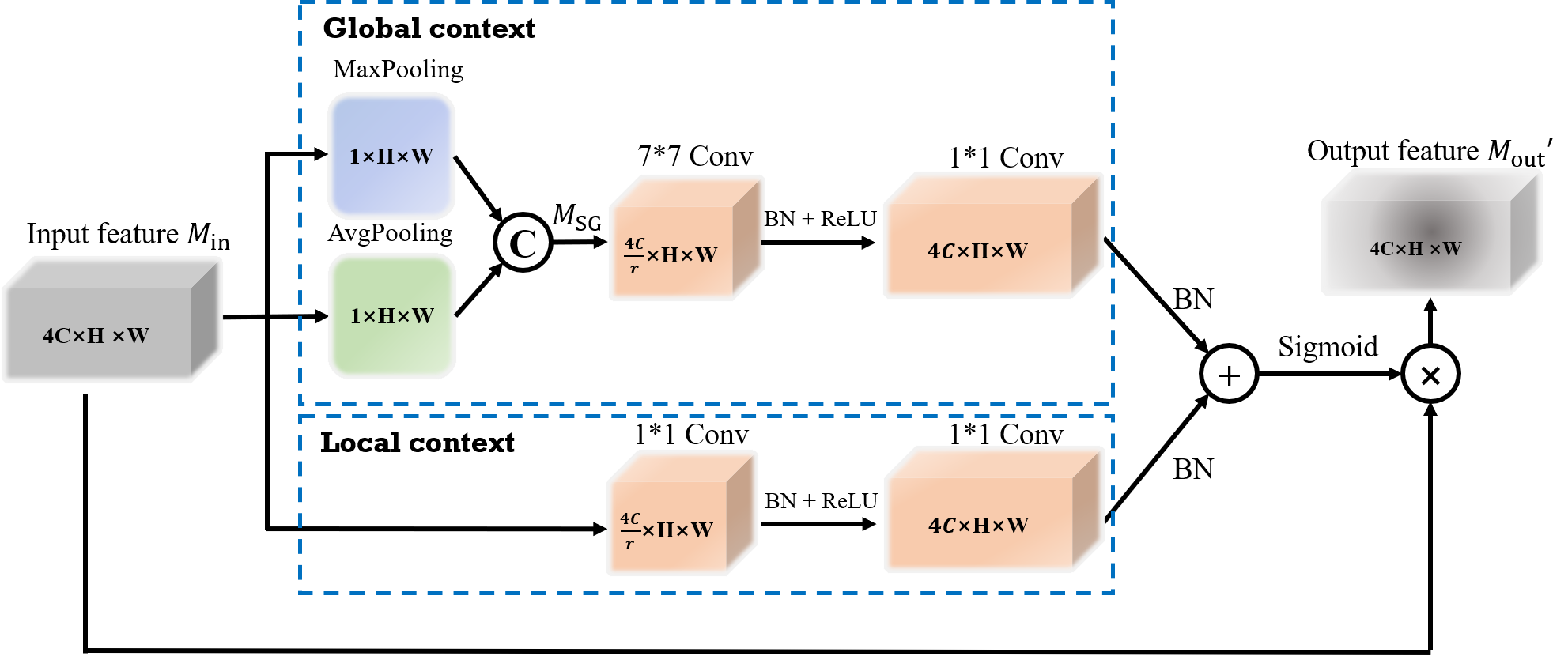}
\end{center}
\vspace{-0.3cm}
   \caption{\textbf{Diagram of the proposed spatial attention module.}}
\label{fig4}
\vspace{-0.5em}
\end{figure}

\textbf{Spatial attention module.} Spatial attention module mainly focuses on the inter-spatial dependencies of the convolutional features and generates spatial attention matrices which highlight informative regions. To calculate the spatial attention maps, we operate global average pooling and max pooling along the channel axis to generate two feature descriptors that indicate max-pooled and average-pooled features. Next, we concatenate the above descriptors and apply convolutional layers to produce the global spatial attention map (see Figure \ref{fig4}). In order to decrease parameter and improve the robustness of training, the first convolution kernel size is set to $\mathbbm{R}$ $^{\frac{4C}{r}\times H\times W}$, where $r$ is the channel dimension reduction ratio.

We add a parallel local branch inside the spatial attention module to enrich feature contexts and improve multi-scale information expression. In this branch, we use the point-wise convolution as the local channel context extractor for each spatial position, among which the convolution kernel size is $1$. Finally, we merge the output feature matrices using broadcasting addition. To emphasize multiple spatial-wise features instead of one-hot activation, we exploit sigmoid function to activate summation. Then we obtain the final spatial weights, which is used to rescale inputs.

\begin{figure}[hp]
\begin{center}
\vspace{-0.3cm}
\includegraphics[width=\linewidth]{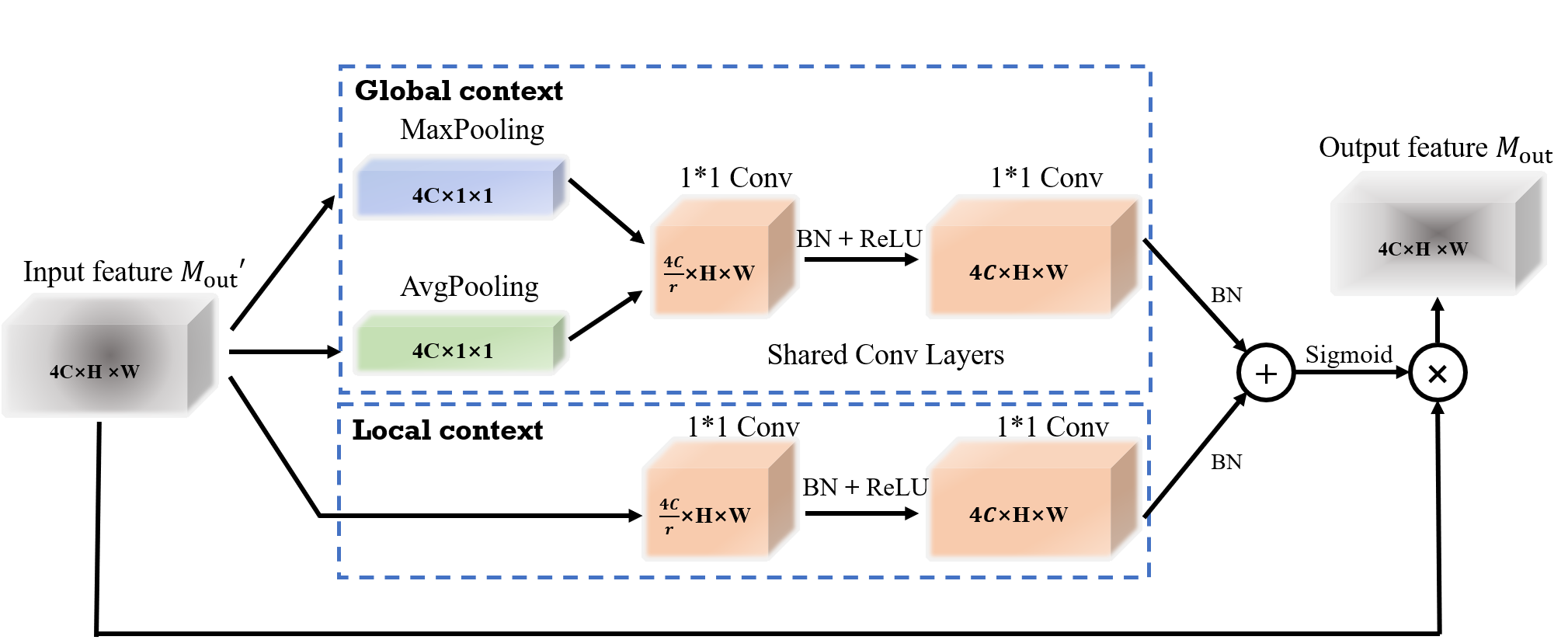}
\end{center}
\vspace{-0.3cm}
   \caption{\textbf{Diagram of the proposed channel attention module.}}
\label{fig5}
\vspace{-0.5em}
\end{figure}

\textbf{Channel attention module.} Different from the spatial attention module, channel attention module is able to capture the inter-dependencies between the channels and learn the inter-channel relationship of features, with the goal of assigning higher weights to the channels with more information. To compute channel attention map effectively, we take the feature map into global spatial module and generate two sets of channel-wise descriptors. As shown in Figure \ref{fig5}, both descriptors are forwarded to a shared multi-layer convolution subnetwork to produce our global channel attention map. The shared subnetwork is composed of two point-wise convolutional layers instead of $fc$ layers.

Similar to the spatial attention, we also inset a parallel local branch into the channel attention module and maintain the same architecture as local spatial attention. Then, we aggregate the output feature maps using broadcasting addition with sigmoid activation function. Spatial weights are used to rescale the input feature map.

\subsection{Mask assembly module} 
For maintaining mask spatial coherence (i.e., pixels close to each other are likely to be part of the same instance) in the feature space, we split the leaf segmentation into two parallel branches: bottom and top branch. The former predicts a set of position-sensitive masks shared by all leaves, since multiple convolutional layers with padding (e.g., with value 0) are inherently translation variant and give the network the ability to distinguish and localize leaves by using different activations. The latter produces corresponding top-level 2D coefficient matrices alongside the box predictions to encode the leaf information. In order to produce original leaf masks, we linearly combine bases according to the mask coefficients. These operations are integrated in mask assembly module.

\begin{figure}[htp]
\begin{center}
\includegraphics[width=\linewidth]{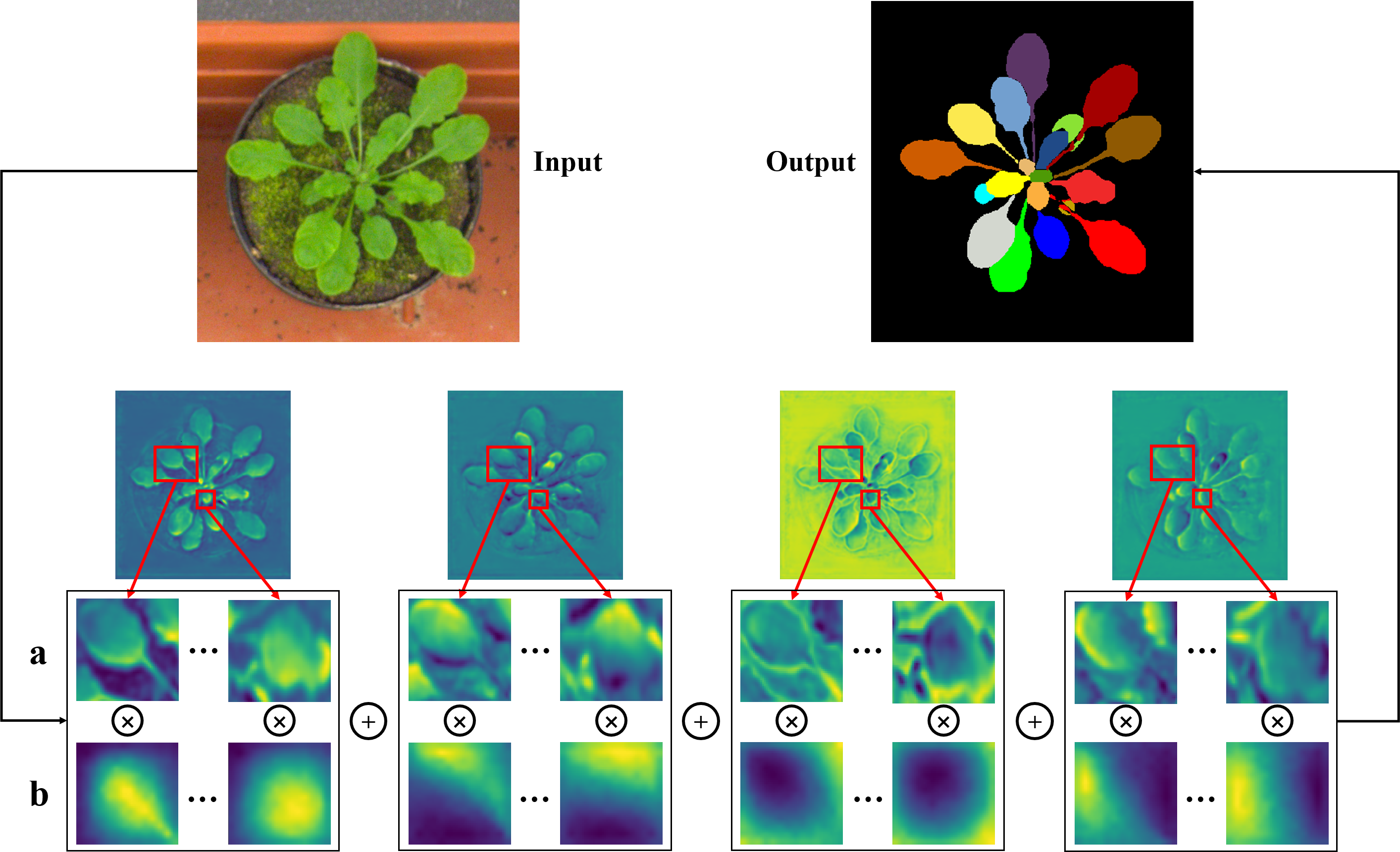}
\end{center}
   \caption{\textbf{Diagram of assembly process.} We illustrate an example of the bases and coefficients. The number of them is $4$ and $76(=4\times 19)$, respectively. 'a' is bases of all leaves and 'b' is the corresponding 2D coefficient matrices. Each base multiplies its coefficient and then is summed to output the final mask.}
\label{fig6}
\end{figure}

The bottom branch is similar to the end-to-end semantic segmentation network, but its weight supervision comes from the mask after assembly. We take ResNet as an encoder to extract more low-level visual characteristics and high-level semantic information. The decoder is comprised of FPN, DAG-Mask, and an upsampling layer, which can output spatial-coherent score maps called bases with the size of $N\times K\times \frac{H}{4}\times \frac{W}{4}$, where $N$ is the batch size and $K$ is the number of score maps.

The top branch follows the object detector FCOS to generate bounding boxes and utilizes NMS to select the best bounding box. In order to learn bases’ coefficients, a single convolution layer is appended on FCOS heads to encode instance-level information including the position and shape of each leaf. Thus, the output of top branch is a set of $2D$ tensors with the number of $N\times K\times P$, where $P$ is the number of bounding boxes.

Given $bases\in\mathbbm{R}^{N\times K\times\frac{H}{4}\times\frac{W}{4}}$, $coefficients\in\mathbbm{R}^{N\times K\times P\times R_C\times R_C}$, and $boxes\in\mathbbm{R}^{P\times 4}$, we first resize them to suitable shapes and then combine them linearly. More specifically, we apply RoIAlign to crop bases with all bounding boxes and resize the region to $R_B\times R_B$. Since the size $R_C$ is smaller than $R_B$, we interpolate coefficients from $R_C$ to $R_B$. The coefficients are illustrated in Figure \ref{fig6} (b). In the end, we utilize element-wise multiplication between each adjusted coefficient and base, and operate summation along $K$ ($K=4$ in Figure \ref{fig6}) dimension to get the original mask.

\subsection{Mask refining module}
After extensive experiments, we notice that leaf boundaries tend to be serrated in masks. Pixels located on the boundaries are hard to be accurately classified because of large ambiguity \cite{li2017not}. Image segmentation is generally operated on regular grids in CNNs, and it will follow the regular sampling pattern between the smooth areas and object boundaries. However, it unnecessarily over-samples the low-frequency regions while concurrently under-samples high-frequency regions. To adaptively select pixels and efficiently computes sharp boundaries, we implement a mask refining module to generate a final anti-aliased and high-resolution mask. The input of the mask refining module is the $P3$ feature vector from FPN and the results from the mask assembly module. Next, we divide the procedure into two parts including point sampling and point prediction.

\textbf{Point Sampling.} During training, we define the set of points to be optimized of each instance as $Set_i$. For each instance, we select the most uncertain points into $Set_i$ according to the output prediction logits from the mask assembly module. To enhance the robustness of the model, we also add some randomly selected points to $Set_i$. Specifically, we define $\beta$ ($\beta >1$) as oversampling rate and $\alpha$ ($0<\alpha \leq 1$) as importance rate. For each instance, we uniformly sample $\beta N$  points from the mask as a set of selectable points $U_i$. We first select the top $\alpha N$ points with logits closest to $0.5$ from $U_i$ and put them into $Set_i$, then randomly select another ($1-\alpha)\times N$ points from the remaining points into $Set_i$. Algorithm \ref{alg:example} provides the pseudo-code of this process.

\begin{figure}[hp]
\begin{center}
\includegraphics[width=\linewidth]{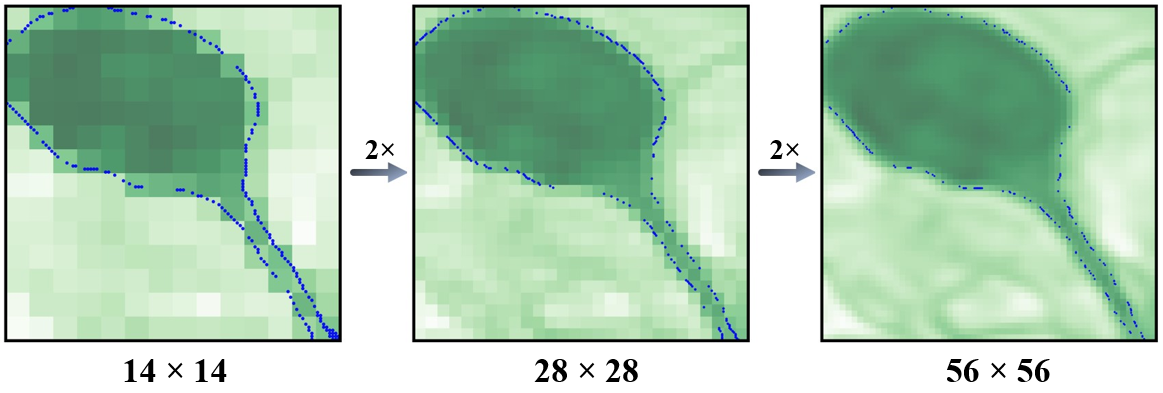}
\end{center}
\vspace{-0.3cm}
   \caption{\textbf{Point sampling strategy.} In each step, we select the most uncertain points in the current mask to update (blue dots shown in the figure), further the $N\times N$ grid is upsampled $2\times$ using bilinear interpolation. This process is repeated until it reaches the targeted resolution.}
\label{fig7}
\vspace{-1em}
\end{figure}

During inference, we use the same point selection strategy for each instance, and select the most uncertain top $N$ points from $U_i$. After reclassifying these most uncertain points, we use bilinear interpolation to amplify the mask and then select the points again. By repeating this process for $x$ times, we can obtain a mask that has been refined step by step (Figure \ref{fig7}).

\textbf{Point prediction.} In order to optimize the logits of the points selected from the above part, we introduce a $n$-layer point-wise convolutional network. The weight of the network in the training phase is shared by all points, that is, it is not affected by the above defined point sampling method. The input of network is the high-precision feature in the $P3$ layer and the output coarse logits from the mask assembly module corresponding to each point.

\begin{algorithm}
   \caption{Point Selection Strategy}
   \label{alg:example}
\begin{algorithmic}
   \STATE {\bfseries Input:} mask logits $M_i$, $S$\\
	\quad\qquad $S$ is either $train$ or $inference$
	
   \STATE $Set_i\gets\{\}$
   \STATE {\bfseries if}\ $S=train$\ {\bfseries then}
    
   \STATE \quad\quad$U\gets uncertainty(M_i)$
   \STATE \quad\quad$P\gets topK(U)$
   \STATE \quad\quad$Set_i\gets Set_i\cup P$

   \STATE {\bfseries else if}\ $S=inference$\ {\bfseries then}
\STATE \quad\quad{\bfseries for}\ $i=1$ {\bfseries to} $step$\ {\bfseries do}

   \STATE \ \quad\quad\quad$M_i\gets interpolation(M_i)$
   \STATE \ \quad\quad\quad$U\gets uncertainty(M_i)$
   \STATE \ \quad\quad\quad$P\gets topK(U)$
   \STATE \ \quad\quad\quad$Set_i\gets Set_i\cup P$

\STATE \quad\quad{\bfseries end for}
   \STATE {\bfseries end if}

\end{algorithmic}
\end{algorithm}

During training time, we define a multi-task loss as:
\begin{align}
\vspace{-1em}
\mathcal{L}=\mathcal{L}_{cls}+\mathcal{L}_{ctr}+\mathcal{L}_{loc}+\mathcal{L}_{mask}+\mathcal{L}_{sem}+\mathcal{L}_{points}
\end{align}
\vspace{-1.3em}
\label{eq9}

where the classification loss $\mathcal{L}_{cls}$, centerness loss $ \mathcal{L}_{ctr}$, location loss $\mathcal{L}_{loc}$ are as same as those in \cite{Tian2019FCOS}. Mask loss $\mathcal{L}_{mask}$ is identical as in \cite{he2017mask}. Semantic segmentation supervision as auxiliary loss with weight $0.3$ is added to GL-DAM mask module and its loss is $\mathcal{L}_{sem}$. $\mathcal{L}_{points}$ is the average binary cross-entropy loss of selected points as defined in point sampling.

\section{Experiments and results}\label{sub4}

\subsection{Implementation details and dataset}

For all experiments, we use a Dell workstation, which is equipped with Intel Xeon (R) CPU $E5-2683V3$ processor and GTX $1080$ Ti GPU. All models are trained on a single $1080$ Ti GPU (take about 1 day). We adopt PyTorch and Detectron$2$ as the deep learning frameworks.

Considering the computational resource and model performance, we set the batch size to $4$. ResNet-101 pre-trained on ImageNet is used as the backbone. We adopt the Kaiming initialization method \cite{he2015delving} with the rectifier’s negative slope of $1$ to initialize other convolution layers. Input images are uniformly normalized to have the shorter side $640$ and longer side at maximum $1440$. Other hyperparameters are set to be the same as FCOS \cite{Tian2019FCOS}.

We use a benchmark dataset of raw and annotated images of plants from LSC of the Computer Vision in Plant Phenotyping and Agriculture (CVPPA) workshop. All the RGB images in the CVPPA LSC belong to rosette plants (Arabidopsis and Tobacco), which are top-view 2D visible-light images from an indoor plant phenotyping platform.

To better evaluate leaf segmentation accuracy, we utilize BestDice metric to estimate the degree of overlap among ground truth and prediction masks. The metric mainly is based on the Dice score of binary segmentation.

\subsection{Ablation experiments}
We carry out a number of ablation experiments to verify the effectiveness of the architectures and hyper-parameters.

\begin{table}[H]
\centering
\caption{\textbf{Comparison of different resolutions and numbers.} Performance with base resolution $28\times 28$ and $56\times56$, coefficient resolution from $7$ to $28$, and number of bases varying from $1$ to $8$. LeafMask uses multi-scale attention module and appends mask refining module. The performance increases as the $R_B$, $R_C$, and $K$ grow, saturating in $56$, $14$, and $4$, respectively.}
\label{table:assembly}
\vskip 0.15in
\begin{center}
\begin{small}
\begin{tabular}{ccccc}
\toprule
$R_B$        & $R_C$         & K & Time(ms)   & Mean\\
\midrule
             & $7\times 7$   & 1& \textbf{123.70}& 78.60 \\
             &               & 2& 126.28& 88.73 \\
$28\times 28$&               & 4& 135.09& 89.68 \\
             &               & 8& 137.86& 89.81 \\
\cline{2-5}
             & $14\times 14$ & 4& 140.77& 89.29 \\
             &               & 8& 140.81& 89.66 \\
\hline
             & \textbf{14}$\times$\textbf{14} & \textbf{4}& 144.69& \textbf{90.09} \\
\textbf{56}$\times$\textbf{56}&     & 8& 145.79& 89.62 \\
\cline{2-5}
             & $28\times 28$ & 4& 162.41& 89.21 \\
\bottomrule
\end{tabular}
\end{small}
\end{center}
\vskip -0.1in
\vspace{-1em}
\end{table}

\textbf{Mask assembly module.} We experimentally measure the performances of the mask assembly module with different $R_C$, $R_B$, and $K$. The combination methods and results are shown in Table \ref{table:assembly}, which shows that: 1) increasing the resolution of coefficient and base can learn more detailed information about leaf instances without introducing much computation; 2)through our experiments, we finally set $R_B=56$, $R_C=14$ and $K=4$ for our baseline model under which the BestDice is optimal and inference time does not increase much; 3)further increasing $R_B$, $R_C$ and $k$ are ineffective mainly because they will clearly increase the training complexity and make it very difficult to predict. Since the bases and corresponding coefficients are linearly combined, it will directly lose the balance and cause poor performance of the network when one base or coefficient matrix is wrong.

\begin{table*}[htp]
\caption{\textbf{Comparison of different attention modules.} Using multi-scale contextual aggregation and dual attention is crucial. The best combination strategy (sequential spatial-channel) further improves the accuracy of leaf segmentation. For all models, we use $R_B=56$, $R_C=14$ and $K=4$ and append mask refining module.}
\label{table:attention}
\vskip 0.15in
\begin{center}
\begin{small}
\begin{tabular}{cccccccc}
\toprule
Attention & Arrangement             & Mean & A1   & A2   & A3   & A4   & A5 \\
\midrule
None      &                         & 88.54& 91.83& 87.32& 88.66& 88.16& 88.40 \\                        
SE        &                         & 88.72& 90.25& 87.22& 90.12& 88.07& 88.71 \\
CBAM      &                         & 88.78& 91.35& 88.40& 88.56& 88.54& 88.67 \\
G-CAM     &                         & 88.82& 90.42& 87.10& 89.31& 88.52& 88.76 \\
GL-CAM    &                         & 89.09& 91.85& 88.34& 90.27& 88.36& 88.97 \\
GL-SAM    &                         & 88.98& 91.64& 87.89& 89.72& 88.42& 88.87 \\
\hline
          & Parallel                & 89.62& 91.64& 89.65& 91.26& 88.82& 89.51 \\
\textbf{GL-DAM}    & Parallel+Shared         & 89.04& 92.05& 88.79& 89.79& 88.42& 88.88 \\
          & Channel-Spatial         & 89.46& 92.33& 88.62& 90.52& 88.76& 89.33 \\
          & \textbf{Spatial-Channel}& \textbf{90.09}& \textbf{92.46}& \textbf{89.66}& \textbf{91.77}& \textbf{89.25}& \textbf{90.00} \\
\bottomrule
\end{tabular}
\end{small}
\end{center}
\vskip -0.1in
\vspace{-1em}
\end{table*}

\textbf{Attention module.} To investigate the effect of multi-scale attention module, we compare the module with SE and CBAM. By contrast, the proposed multi-scale attention module surpasses the other methods (see Table \ref{table:attention}). One crucial reason is that global attention modules lack the local context information, which aggravates the problems brought by the scale variation of leaf. We construct seven attention modules to select the optimal architecture. Table \ref{table:attention} presents the experimental results on the LSC dataset. It can be seen that utilizing multi-scale contextual aggregation (global and local information) to generate attention maps outperforms single attention, and the spatial-first order is the best arrangement mode.

\begin{table}[H]
\centering
\caption{\textbf{Comparison of different $\beta$ and $\alpha$.} Performance of LeafMask with different oversampling rate $\beta$ and importance rate $\alpha$. The last column indicates the model performance without the mask refining module.}
\label{tab5}
\vskip 0.15in
\begin{center}
\begin{small}
\begin{tabular}{cccc}
\toprule
Point refining &$\beta$        & $\alpha$         & Mean\\
\midrule
$\checkmark$& 1          & 0.5           & 89.47 \\
$\checkmark$& \textbf{3} & \textbf{0.75} & \textbf{90.09} \\
$\checkmark$& 5          & 1             & 89.21 \\
\hline
-      & -          & -             & 89.02 \\
\bottomrule
\end{tabular}
\end{small}
\end{center}
\vskip -0.1in
\vspace{-1em}
\end{table}

\textbf{Mask refining module.} We compare the performance of the model under different $\beta$ and $\alpha$, and the model performs best with $\beta=3$ and $\alpha=0.75$. When the value of $\beta$ is small, the selected points tend to be uniformly distributed, so their logits are also uniformly distributed between $0-1$. The feature space that $n$-layer network needs to learn is relatively large, which possibly exceeds its learning ability, so the model does not perform well. On the contrary, when the value of $\beta$ is high, the logits of the selected points is concentrated around $0.5$, and the $n$-layer network lacks other points’ information to accurately classify these points, so it will also get poor performance.

\subsection{Comparison with state-of-the-art methods}
\begin{figure*}
\begin{center}
\includegraphics[width=0.9\linewidth]{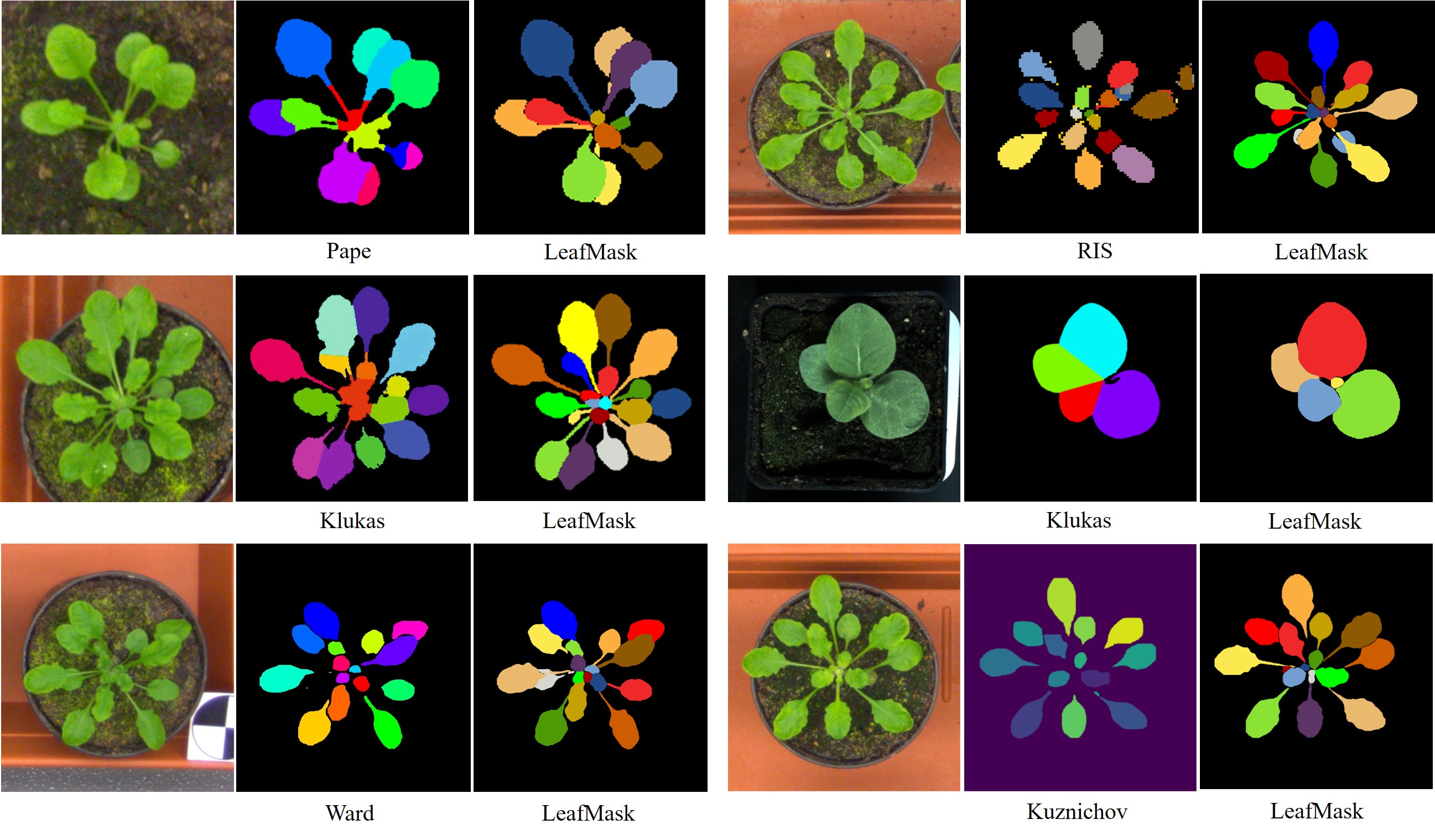}
\end{center}
\vspace{-1em}
   \caption{\textbf{Qualitative results with main-stream leaf segmentation methods.} In each set of comparisons, the middle is the predicted mask provided by the original paper (Pape\cite{pape2015utilizing}, RIS\cite{romera2016recurrent}, Klukas\cite{pape20143}, Ward\cite{ward2018deep}, Kuznichov\cite{kuznichov2019data}), and the right is the corresponding results of Leafmask.}
\label{fig8}
\vspace{-1em}
\end{figure*}

\begin{figure*}
\begin{center}
\includegraphics[width=0.9\linewidth]{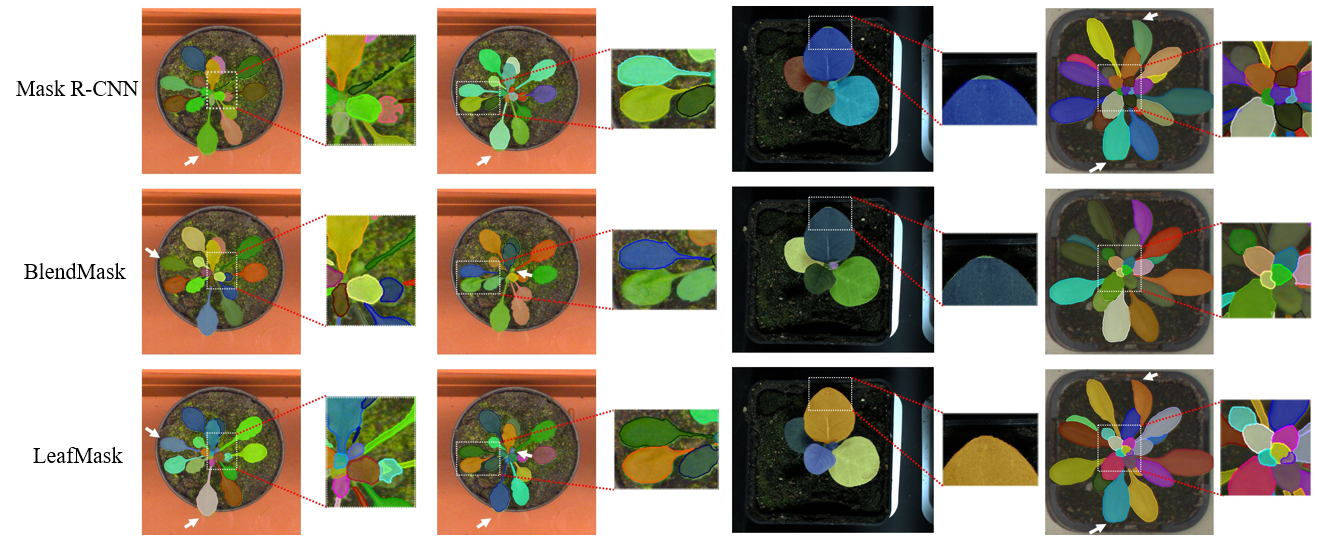}
\end{center}
\vspace{-1em}
   \caption{\textbf{Qualitative results with other instance segmentation approaches.} To make a fair comparison with Mask R-CNN, BlendMask, and LeafMask, the code base we use for qualitative results is \texttt{Detectron2}. Recently released \texttt{Detectron2} originates from \texttt{maskrcnn\_benchmark} with significant enhancements for performance.}
\label{fig9}
\vspace{-1em}
\end{figure*}

We compare LeafMask with other state-of-the-art algorithms on LSC dataset. As shown in Table \ref{table:cvppp_results}, our proposed model achieves $90.09\%$ mean BestDice score for all test sets, which is better than other approaches. Our LeafMask also outperforms the Mask R-CNN and BlendMask with the same backbone-neck ResNet-101-FPN by $3.2\%$ and $2.2\%$ in BestDice score, respectively. Note that most works report results on A1 subset only, and the A4,A5 subsets are added later. The qualitative results of our LeafMask are also shown in Figure \ref{fig8} and Figure \ref{fig9}.

\begin{table}[H]
\centering
\caption{\textbf{CVPPA LSC results.} Segmentation performance comparison (BestDice). Statistical evaluation results provided by the Leaf Segmentation Challenge board, based on the submitted image analysis results for the testing-dataset.}
\label{table:cvppp_results}
\vskip 0.15in
\begin{center}
\begin{small}
\setlength{\tabcolsep}{1.8mm}
\begin{tabular}{ccccccc}
\toprule
Method        & Mean         & A1           & A2           & A3             & A4           & A5    \\
\midrule
RIS+CRF\cite{romera2016recurrent}       & -            & 66.6         & -            & -              & -            & -     \\
MSU\cite{scharr2016leaf}           & -            & 66.7         & 66.6         & 59.2           & -            & -     \\
Nottingham\cite{scharr2016leaf}    & -            & 68.3         & 71.3         & 51.6           & -            & -     \\
Wageningen\cite{yin2014multi}    & -            & 71.1         & 75.7         & 57.6           & -            & -     \\
IPK\cite{pape20143}           & 62.6         & 74.4         & 76.9         & 53.3           & -            & -     \\
Pape\cite{pape2015utilizing}          & 71.3         & 80.9         & 78.6            & 64.5              & -            & -     \\
Salvador\cite{salvador2017recurrent} & - & 74.7 & - & - & - & - \\
Brabandere\cite{de2017semantic} & - & 84.2 & - & - & - & - \\
Ren\cite{ren2017end}           & -            & 84.9         & -            & -              & -            & -     \\
Zhu\cite{zhu2018joaquin}           & -            & -            & -            & -              & 87.9         & -     \\
Ward\cite{ward2018deep}          & 81.0         & 90.0         & 81.0         & 51.0           & 88.0         & 82.0          \\
Kuznichov\cite{kuznichov2019data}     & 86.7         & 88.7         & 84.8         & 83.3           & 88.6         & 85.9  \\
Mask R-CNN      & 86.9             & 89.4             & 85.0             & 87.6          & 86.4             & 86.8 \\
BlendMask     & 87.9        & 90.1        & 87.4        & 89.8          & 87.1        & 87.9  \\
\textbf{LeafMask} & \textbf{90.1}& \textbf{92.5}& \textbf{89.7}& \textbf{91.8}  & \textbf{89.3}& \textbf{90.0} \\
\bottomrule
\end{tabular}
\end{small}
\end{center}
\vskip -0.1in
\vspace{-1.5em}
\end{table}

Among the main-stream leaf segmentation methods (see Figure \ref{fig8}), it is difficult for Pape and klukas to accurately segment two or more overlapping blades, and the boundary of the two leaves is often mixed. For Ward and Kuznichov, it’s hard to detect and segment small blades that the small blades are often missed. The RIS segmentation of the leaf boundary tends to have a more obvious jagged mask and other methods also cause different degrees of segmentation loss on the boundary of the instance, especially for the petiole. Our algorithm has solved the above problems well. As shown in Figure \ref{fig9}, compared with Mask R-CNN and BlendMask, LeafMask can segment the arc-shaped boundary of the blade well. What's more, small leaves will not be missed in dense regions of petioles or small leaves.

\section{Conclusion}
We have proposed a new end-to-end anchor-free one-stage instance segmentation model towards greater accuracy on leaf segmentation. By inserting multi-scale attention module into DAG-Mask branch, combining assembly module with coefficient predictor, and adding mask refining module, LeafMask have achieved $90.1\%$ best dice and outperformed all state-of-the-art approaches on CVPPA LSC test sets. We hope that our LeafMask will serve as a baseline to motivate further investigation of leaf segmentation for various plant phenotyping tasks.

\section{Acknowledgments} 
The work is supported by National Key R$\&$D Program of China (2020YFD0900204) and Key-Area Research and Development Program of Guangdong Province China (2020B0202010009).

{\small
\bibliographystyle{ieee_fullname}
\bibliography{citation}
}

\end{document}